\documentclass[11pt,a4paper]{article}
\usepackage[hyperref]{acl2021}
\usepackage{times}
\usepackage{latexsym}

\usepackage{graphicx}
\usepackage{amsmath}
\usepackage{amssymb}
\usepackage{epsfig}
\usepackage{multirow}

\usepackage{microtype}

\aclfinalcopy 

\title{Ensemble-based Transfer Learning for Low-resource Machine Translation Quality Estimation}

\author{Ting-Wei Wu \\ Georgia Institute of Technology \\ \texttt{waynewu@gatech.edu} \\
        \And
        Yung-An Hsieh  \\ Georgia Institute of Technology \\ \texttt{yhsieh37@gatech.edu} \\
        \AND
        Yi-Chieh Liu  \\ Georgia Institute of Technology \\ \texttt{yliu3233@gatech.edu} \\}

\date{}

\begin{document}
\maketitle
\begin{abstract}
Quality Estimation (QE) of Machine Translation (MT) is a task to estimate the quality scores for given translation outputs from an unknown MT system. However, QE scores for low-resource languages are usually intractable and hard to collect. In this paper, we focus on the Sentence-Level QE Shared Task of the Fifth Conference on Machine Translation (WMT20), but in a more challenging setting. We aim to predict QE scores of given translation outputs when barely none of QE scores of that paired languages are given during training. We propose an ensemble-based predictor-estimator QE model with transfer learning to overcome such QE data scarcity challenge by leveraging QE scores from other miscellaneous languages and translation results of targeted languages. Based on the evaluation results, we provide a detailed analysis of how each of our extension affects QE models on the reliability and the generalization ability to perform transfer learning under multilingual tasks. Finally, we achieve the best performance on the ensemble model combining the models pretrained by individual languages as well as different levels of parallel trained corpus with a Pearson's correlation of 0.298, which is 2.54 times higher than baselines.
\end{abstract}

\section{Introduction}

Quality Estimation (QE) of Machine Translation (MT) is a task to estimate the quality scores for given translation outputs from an unknown MT system at run-time, without relying on reference translations \cite{blatz2004confidence,specia2009estimating}. QE can be performed at various granularity (sentence/word/phrase) levels. We focus on sentence-level QE, which aims at predicting how much post-editing effort is needed to fix the translations \cite{specia2018quality}.

The common methods regard the sentence level QE as a supervised regression task, which uses quality-annotated noisy parallel corpora, specifically, QE data, as training data \cite{specia2018quality}.
Previous studies on QE \cite{felice2012linguistic,gonzalez2012prhlt,specia2013quest,shah2015bayesian} generally separate the model into two independent modules: feature extractor and machine learning module. 
The feature extractor is designed to extract human-readable features that describe the translation quality, such as source fluency and translation complexity. 
Based on extracted features, the machine learning module then predicts how much effort is needed to post-edit translations to acceptable results as measured by the Human targeted Translation Edit Rate (HTER) \cite{snover2006study}. 
As the great success of deep neural network (DNN) in the natural language processing (NLP) realm, researchers have applied DNN to QE tasks and achieve significant improvement \cite{kreutzer2015quality,kim2016recurrent,kim2017predictor,kepler2019unbabel,fan2019bilingual}. 
In this study, our proposed sentence-level QE model is extended from the Predictor-Estimator model \cite{kim2017predictor}, which consists of a neural word prediction model (predictor) and a neural QE model (estimator).

As with most supervised machine learning problems, an important issue in QE models is the need for labeled data. Unfortunately, QE data are expensive and not readily available because QE annotations are based on human post-edits, which require revision of the originally translated sentence \cite{specia2018quality}. Therefore, in addition to advancing the state of the art at sentence level, our more specific goals also include to investigate the following:
\begin{itemize}
    \item The effect of training data size imposed on the training process and scores.
    \item The reliability of QE model structures under multilingual tasks with few data samples.
    \item The generalization ability of QE models to perform transfer learning on different languages.
\end{itemize}

We mainly adopt multiple language resources for pretraining a predictor that encapsulates the semantic knowledge agnostic of languages; then retrain the predictor-estimator for generating QE scores in low-resource languages. Finally, we adopt two ensembling techniques to incorporate different aspects of trained models for better quality estimation. Our contributions are as follows:
\begin{itemize}
  \item We explore a new task to estimate QE scores of translation outputs when QE scores are almost unseen in that particular language pair during training. We extend BiRNN \cite{schuster1997bidirectional} with Transformer \cite{vaswani2017attention} as predictor in QE model and introduce new losses: noise-contrastive estimation (NCE) and negative sampling loss (NEG) for predictor training.
  \item We apply transfer learning to fine-tune predictors, which are pretrained with individual language pairs, with other multiple languages. The better scores implicate the knowledge learned from pretrained weights is useful in our downstream task cases.
  \item We propose new ensembling techniques for combining the models pretrained by single language as well as different levels of parallel trained corpus, then fine-tuned by all languages. Finally we achieve the best performance of Pearson correlations score \textbf{0.298} on the final ensembling model, which is \textbf{2.54} times than the baseline model we have.
\end{itemize}

\section{Problem Statement}
Our study follows the schema of Sentence-Level QE Shared Task of the Fifth Conference on Machine Translation (WMT20) \cite{barrault-etal-2019-findings}, including the data, baseline, and evaluation metric, but in a more challenging setting. 
Given a translation result $(x, y)$ from an unknown MT system translating from language A to B where $x=\{x_1, x_2, ..., x_{T_x}\}$ is the source sentence; $y=\{y_1, y_2, ..., y_{T_y}\}$ is the translated sentence, we aim to predict a QE score $s$. $T_x$, $T_y$ denotes length of the source and target sentences. However, in training phase, we are only given few QE pairs $(x^{(i)},y^{(i)},s^{(i)})$ of targeted translation languages (A, B). This section will focus on the data and metric of the shared task and we will provide detailed introduction of the baseline in the next section.

\begin{table}[t]
  \caption{Data statistics}
  \label{tab:data}
  \footnotesize
  \centering
  \begin{tabular}{l||c|c|c|c}
    \hline
    \textbf{Dataset} & \textbf{Datatype} & \textbf{train} & \textbf{valid} & \textbf{test}      \\
    \hline\hline
    ro-en & QE & 7K & 1K & 1K \\
    \hline
    et-en & QE &  7K & 1K & 1K \\
    \hline
    ne-en & QE &  7K & 1K & 1K \\
    \hline
    en-de & QE &  0.1K & 0.1K & 1K \\
    \hline
    en-zh & QE &  0.1K & 0.1K & 1K \\
    \hline\hline
    en-de & Parallel &  10/20/30/50 K & - & - \\
    \hline
    en-zh & Parallel &  10/20/30/50 K & - & - \\
    \hline
  \end{tabular}
\end{table}

\begin{figure*}[t]
    \center
    \includegraphics[width=1.0\textwidth]{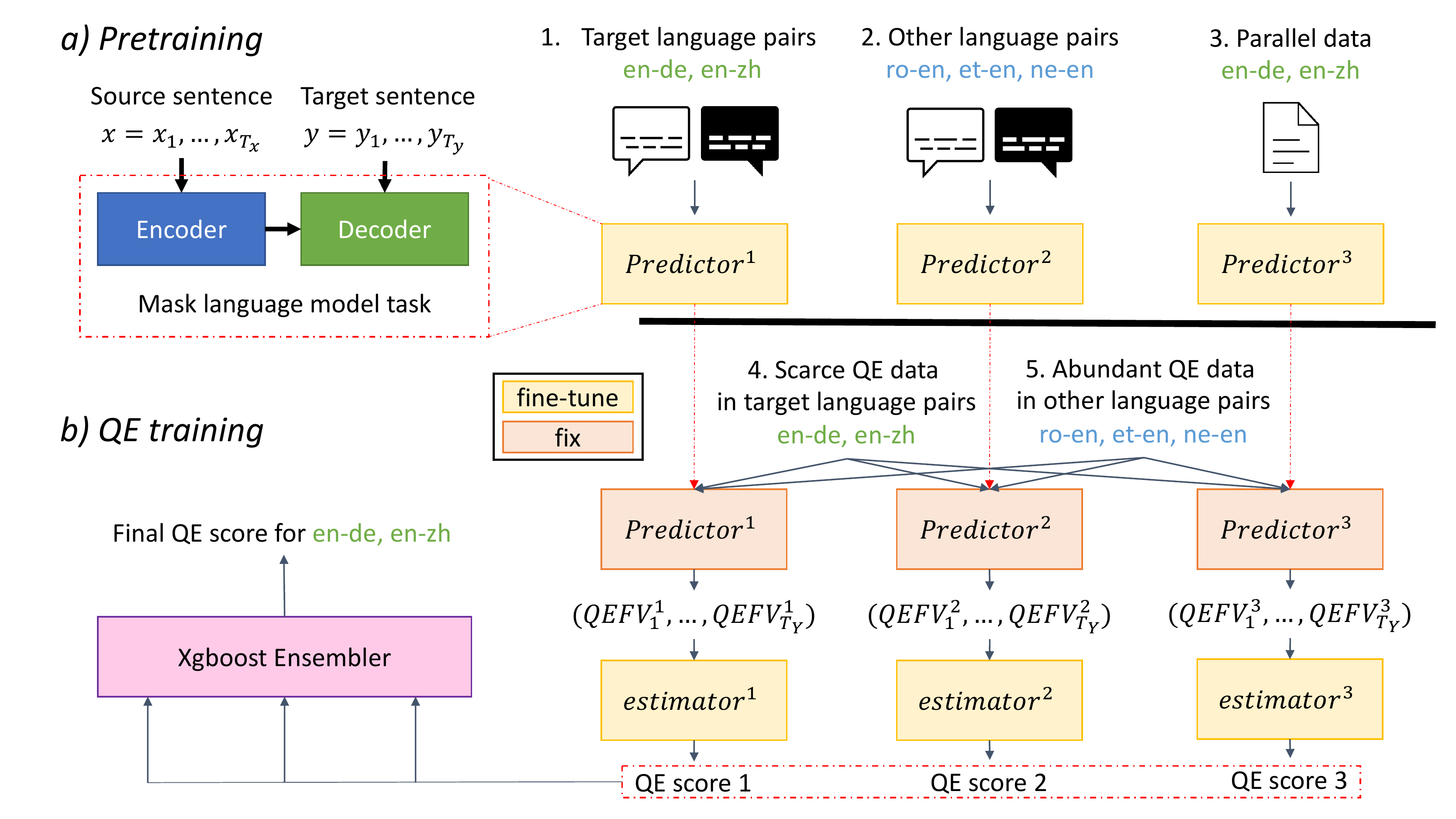}
    \caption{Our proposed framework architecture. a) We first pretrain multiple predictors with the mask language model task on three different kinds of translation corpus. b) We use pretrained predictors to extract QEFV vectors and fine-tune estimators with QE data; finally we use ensemble model to prodcue final QE scores.}
    \label{fig:arch}
\end{figure*}

\subsection{Data}
The QE data we use consist of 5 language pairs, which includes English-German (en-de), English-Chinese (en-zh), Romanian-English (ro-en), Estonian-English (et-en), and Nepalese-English (ne-en), as shown in Table. \ref{tab:data} (4. and 5. in Figure.\ref{fig:arch}). To simulate the low-resource setting, we only use 100 sentences in en-de and en-zh for training and test our model's performance of transfer learning from other language pairs.

The data were collected by translating sentences sampled from source language Wikipedia articles using state-of-the-art neural machine translation (NMT) models \cite{ott2019fairseq}. Different from previous QE tasks, which labeled the translation quality based on post-editing, each translation was annotated with Direct Assessment (DA) scores by at least three professional translators in WMT20 \cite{barrault-etal-2019-findings}. The DA score ranges from 0-100, as higher the better. The implemented QE models aim at predicting the average of z-standardized DA scores of each translation.

We also leverage three additional sources for predictor pretraining. First we obtain the pretrained weights of predictors from (1.) target and (2.) other language pairs as shown in Figure. \ref{fig:arch}. We also extract additional parallel data of en-de and en-zh language pairs to enrich model's efficacy from the News-Commentary parallel corpus \cite{TIEDEMANN12.463}.

\subsection{Metric}
In this study, QE models are evaluated in terms of the Pearson's correlation metric for the predicted DA scores against human DA scores on the development data sets. We evaluate the QE models on both multilingual basis and per-language basis. 

\section{Approach}

We adapts "OpenKiwi" \cite{kepler2019openkiwi}, an open-source framework for QE task, to construct our proposed ensemble-based QE model in Figure \ref{fig:arch}. Similar as other state-of-the-art methods \cite{kim2017predictor, wang2018alibaba}, we use a neural-based architecture, which is mainly based on the predictor-estimator architecture initially proposed from \cite{kim2017predictor}. As shown in Figure \ref{fig:arch}, our proposed QE model architecture involves an ensemble regression model based on XGBoost \cite{xgboost} to combine all prediction results from several QE models, each of which contains a feature predictor and a quality estimator. 

\subsection{Predictor-Estimator}


Our baseline model follows the implementation described in \cite{kim2017predictor}, which applies a bi-directional RNN encoder-decoder architecture as a predictor. 
The mask language modeling (LM) \cite{devlin2018bert} serves as the task to train the predictor, in which the predictor predicts a masked target word $y_j$ conditioned on the source context $x$, and target context $y_{-j}=(y_1, y_{j-1}, y_{j+1}..., y_{T_{y}})$ and acquires QE feature vectors (QEFVs). We illustrate it in the equation \ref{eq:prob_model}.
\begin{align}
    P(y_j|y_{-j},x)&=g([\overrightarrow{s}_{j-1};\overleftarrow{s}_{j+1}],[y_{j-1};y_{j+1}],c_j)\\
    &=\frac{exp(w_{j}^{T}W_{s_{j}})}{\Sigma_{k=1}^{K_y}exp(w_{k}^{T}W s_{j})} 
    \label{eq:prob_model}
\end{align}
where $g$ is a nonlinear function. $s_j = [\overrightarrow{s}_{j-1};\overleftarrow{s}_{j+1}]$ refer to the concatenation of $\overrightarrow{s}_{j-1}$ and $\overleftarrow{s}_{j+1}$, which are the hidden state at the last layer of forward decoder and backward decoder of target sentence respectively. We denote $c_j$ as the source context vector resulting from the attention mechanism when predicting the current target word. $w_j \in \mathbb{R}^{K_y}$ is denoted as one-hot representation of target word and $W \in \mathbb{R}^{2d \times K_y}$ is the weight matrix. After training the model, QEFV could be calculated as follows:
\begin{align}
    QEFV_j = [(w_j^T W \odot s_j^T)]^T
    \label{eq:qefv}
\end{align}
The quality estimator is based on Bidirectional Long Short-term Memory (BiLSTM) for predicting quality scores using QEFVs as inputs.
\begin{align}
    Q_{Estimator}(y, x) = \sigma(Wh_{src:tgt}))
    \label{eq:estimator}
\end{align}
In the calculation, $h_{src:tgt}$ represents the average of $BiLSTM(QEFVs)$. In general, the model solves the supervised regression task that aims at estimating the HTER score of a target translation.

In addition, we implement the transformer based predictor. To modify baseline predictor, we replace the encoder and decoder in the predictor with the transformer module \cite{vaswani2017attention}. 


The attention in transformer is computed on a set of queries simultaneously, packed together
into a matrix $Q$. The keys and values are also packed together into matrices $K$ and $V$. The attention is described as follows:
\begin{align}
    Attention(Q,K,V)=softmax(\frac{QK^T}{\sqrt{d_k}})V
    \label{eq:attention}
\end{align}
$d_k$ is denoted as the dimension of K. We compare the results of transformer with the baselines.


\subsection{Loss}
\label{sec:loss}
Since the predictor is trained to predict the token of the target sentence given the source and the left and right context of the target sentence, we mainly adopt two different loss functions that are commonly used in masked language model settings. The first one is Cross Entropy Loss between the multinomial distribution $P_j$ predicted from the model and the true token distribution.
\begin{equation}
    Cross Entropy Loss=-\sum_{i}^{N} \sum_{j}^{T}y^{(i)}_{j}log(P^{(i)}_{j})
\label{eq:ce}
\end{equation}

where $N$ is the number of data, $T$ is the time step and $y_j$ is the one-hot vector with the length of vocabulary size $|V|$ representing the true token id. It is equivalent to maximize the likelihood of target sequences given the predicted multinomial distribution $P$. It is also used in Predictor-Estimator structure in our baseline model from Openkiwi \cite{kepler2019openkiwi}.

We also introduce another kind of losses: Noise-Contrastive Estimation (NCE) and Negative Sampling Loss (NEG) into our optimization setting. The Loss function for NCE is in Equation \ref{eq:nce}.

\begin{align}
    J(\theta) &= -(log(\frac{exp(Wx_{w_i})}{exp(Wx_{w_i}+k\cdot Q(w))}) \\
                         &+ \sum_{w^{'}_{i} \notin w_{i}}^{k} log(1-\frac{exp(Wx^{'}_{w_i})}{exp(Wx^{'}_{w_i}+k\cdot Q(w))})
\label{eq:nce}
\end{align}

Noise-contrastive estimation (NCE) is a sampling loss typically used to train classifiers with a large output vocabulary space. Updating the softmax gradient over a large number of possible classes is prohibitively expensive. Therefore, using NCE, we can reduce the problem to binary classification problem by training the classifier to discriminate between samples from the “real” distribution and an artificially generated noise distribution $Q(w)$. $x_{w_i}$ here represents the hidden states we have obtained from the predictor; then we use $W$ to map them into the output vocabulary space. But instead of updating entire vocabulary size as in normalization constant in softmax; we only sample $k$ negative words $w_i^{'}$ to update the gradients. Also Negative sampling loss (NEG) is a simple version of NCE used in word2vec \cite{word2vec}. The equation is simplified as the following:

\begin{align}
    J(\theta) = -(log(\sigma(Wx_{w_i})) + \sum_{w^{'}_{i} \notin w_{i}}^{k} log(-\sigma(Wx^{'}_{w_i}))
\label{eq:neg}
\end{align}

Here we still extract $k$ negative words $w_i^{'}$ to update the gradients; however, we directly treat the logits with sigmoid function directly to output the probability of it being to the real target.

\subsection{Ensembling}
\label{sec:ensemble}
To further leverage the benefits of exploiting models trained on multiple languages and data, ensembling techniques are applied to our QE model. We utilize a novel strategy consisting of learning a convex combination of system predictions along with weights learned on the development sets. Namely, we have models outputing sentence-level predictions directly on development sets separately. Then we treat them as additional features along with the original source and target tokens. Our goal is to learn a linear combination of these features using a regression model directly optimizing task scores on development sets. To prevent overfitting, we also train the regression model with $k$-fold cross validation. For regression model choice, we mainly adopt \textbf{ridge regression model} \cite{ridge} and \textbf{XGBoost model} \cite{xgboost} for our training media. 

Ridge regression treats the task as a linear regression problem with penalty over the weights $w$ with a regularization constant $\alpha$ as shown in Equation \ref{eq:ridgeloss}. It aims to minimize the mean square error between prediction scores and real scores while preventing the over growth from the model size. 
\begin{equation}
    L(y,\hat{y} ; w) = ||y - \hat{y}||^2_2 + \alpha * ||w||^2_2
    \label{eq:ridgeloss}
\end{equation}

On the other hand, XGBoost algorithm is an extension of gradient boosting tree algorithm which is popular in ensemble learning. The essential concept is to learn multiple sub regressors trained on residuals between true scores and the prediction of the previous weak regression model; then perform gradient descent on the loss between prediction summation of these models and the true scores. Therefore, the objective function at time step $t$ will be shown in Equation \ref{eq:xgboostloss}, where $g_i = \partial_{\hat{y_i}^{(t-1)}} l(y_i,\hat{y_i}^{(t-1)}) $ and $h_i = \partial^2_{\hat{y_i}^{(t-1)}} l(y_i,\hat{y_i}^{(t-1)})$ serving as first and second derivative terms of Taylor approximation for loss function and $f_t(x_i)$ will be the prediction of the current regression model at time step $t$ on $x_i$. Here we could define any form of loss function $l(y_i,\hat{y_i}^{(t-1)})$ to optimize the final $L(y,\hat{y}^{(t-1)})^{(t)}$. And $\Omega(f_t)$ serves as the model complexity term where we could define hyperparameters to regularize the model.
\begin{align} 
 L(y,\hat{y}^{(t-1)})^{(t)} = \sum_{i=1}^n [l(y_i,\hat{y_i}^{(t-1)}) + g_i f_t(x_i) \notag \\
                                 + \frac{1}{2} h_i f^2_t(x_i)] +\Omega(f_t) 
\label{eq:xgboostloss}
\end{align}

\section{Experiment Results}
The experimental results of our approaches for WMT20 sentence-level QE shared task are shown in Table \ref{tab:results}. 

\subsection{Experimental Setting}
In our baseline predictor, we adopt two LSTM layers and hidden state size equals to 400. The dropout rate is 0.5. For the proposed transformer-based predictor, the number of layers for the self-attention encode and forward/backward self-attention decoder are all set as 6, where we use 4-head self-attention in practice. The dropout rate is set as 0.1.
For the quality estimator module, we set one LSTM layers and a hidden state with size 400. The dropout rate equals zero. The estimator model we use for all the proposed predictor models holds the same parameter. We uniformly trained all the described models with Adam \cite{kingma2014adam} with 2e-3 learning rate and batch-size which equals 64.
We furthermore investigate the task from five perspectives and design the experiments, which are the model structure, training loss, pretrained predictor, addition of parallel data, and model ensembling.

\subsection{Results of Baseline}
\label{sec:baseline}
As shown in Table \ref{tab:results}, our implementation of the baseline model is slightly lower than the baseline provided by WMT20 QE shared task. However, our performance tendency for different language pairs aligns with it. The correspondence ensures our baseline results could serve as a valid baseline to observe the effect on our other experiments.

Although we trained our baseline with the same amount of data for each language pairs, we observed that the performances on et-en, ne-en, and ro-en pairs were much better than en-de and en-zh pairs. The possible reason is the difference between the target languages and the fact that the QEFVs from the predictor are extracted from the hidden states of target sentences. English serves as the target language for all et-an, ne-en, and ro-en pairs, which provides more training instances to make the trained predictor and QEFVs more representative for English. 
On the other hand, en-de and en-zh pairs have their specific target languages (German and Chinese). The lack of training instances for these two target languages can result in worse performance than the other three language pairs.

\subsection{Results of Transformer Predictor}
\label{sec:transformer}
With the replacement of BiLSTM architecture in predictor using Transformer, we didn't achieve better performance than the baseline. The performance on most of the language pairs had lower Pearson's correlation. We concluded two reasons that may lead to this result. First, we observed that when training the transformer-based predictor, more fine-tuning processes, including more iterations and lower learning rate, were required compared to training the BiLSTM predictor. 
Second, the architecture of our transformer-based predictor may requires further adjustment according to \cite{fan2019bilingual}. Instead of using the mismatch features as described in \cite{fan2019bilingual}, our implementation solely rely on QEFVs for contextualized representations. Further investigation on the architecture and fine-tuning process of the transformer-based predictor can be conducted to achieve better performance.

\subsection{Results of Applying Different Loss}
\label{sec:losses}
Our second variations is the use of NCE and NEG losses as mentioned in section \ref{sec:loss}. Here we define $100$ negative words to sample from the distribution built from vocabulary frequency. Given the nature of NCE and NEG loss, they both tend to ameliorate the complexity efforts of updating softmax gradient from a given large vocabulary size. Therefore, such introduction of negative word randomness should expectedly result in faster convergence by succinctly adding the correct gradients. However, the results shown in Table \ref{tab:results} do not explicitly reveal such advantages and are slightly poorer compared with the baseline model. 

We mainly concluded two reasons that may conduct such negativeness. One is that shrinking the gradient space to update may result in harder convergence for the optimizer since weights for most of the vocabulary are not changed throughout several updates. While it is useful in Word2vec setting when updating correct gradients, it is inferior in our case especially when it comes to various different languages. The data in our vocabulary space contain obvious different nature from different languages, which indicate some of vocabulary would not be visited during the certain amount of the training process. The final embeddings extracted from those words may not contain rich semantic information from the original sentences; on the other hand, it gives worse representation without gradient updating based on every data instance. Second, our vocabulary size may not be large enough to exploit the advantage of using negative sampling loss, where every word in our case may disclose essential information to update other words in vocabulary even if they don't appear in the sentence. In lack of updating their corresponding  gradients, training data could not be fully exploited in better representation learning, or even worse. 

\subsection{Results of Fine-tuning pretrained Predictor}
\label{sec:pretraining}
Next, we experimented on the effect of predictors pretrained with different language pairs by using the trained predictor weights provided along with the WMT20 shared task and OpenKiwi \cite{kepler2019openkiwi}. We utilized the weight except for the embedding layers. Each pretrained predictor of a specific language pair was fine-tuned with the QE data of all five language pairs used in this study. As shown in Table \ref{tab:results}, using predictor pretrained with en-de and ro-en pairs resulted in better performance than the baseline The ro-en pretrained predictor achieved the highest Pearson score of 0.151 for the de+zh testing set.

An interesting observation is that using the predictor pretrained on a specific language pair didn't lead to improvement on that language but instead improve other languages. For example, using predictor pretrained with en-de pair resulted in a lower score on en-de pair but improved the performance of en-zh, ro-en, and ne-en pairs. 
A possible explanation is that we only used the weights of the pretrained model but not the words embedding. 
This leads to a mismatch between the vocabulary used to pretrain the models and the vocabulary used to fine-tune the models for the QE task.
Still, the improvement achieved by pretraining the predictor demonstrates the generalization ability of the QE model to perform transfer learning on different languages. 

\subsection{Results of Adding Parallel Data}
\label{sec:parallel}
In this experiment, we added different amounts of en-de and en-zh parallel data from News-Commentary parallel corpus when training the predictor. More specifically, we added 10K, 20K, 30K, and 50K of data for each language pair, which are indicated as experiment $D1$, $D2$, $D3$, and $D5$ respectively in Table \ref{tab:results}. From the results, we observed that additional parallel data for training predictor led to better performance. However, adding more data didn't always lead to further improvement. As shown in Table \ref{tab:results}, we achieved the best performance in $D2$ with a Pearson score of 0.146 for de+zh pairs and 0.473 for et+ne+ro pairs. As we continuously added more data, the performances continuous to drop for en-de and en-zh pairs. We concluded that this is due to larger vocabulary being built as we utilized more parallel data, which further led to sparsity as we trained the estimator model using only the 7K training data from WMT20 and caused the degradation of performance.
We also observed that although we added parallel data for only en-de and en-zh pairs, the performances of other three language pairs (et-en, ne-en, and ro-en) also improved. This indicates that the model can transfer the knowledge from a language to others. 

\subsection{Results of Ensembles}
\label{sec:ensemble_results}
As depicted in section \ref{sec:ensemble}, we utilized two strategies for ensembling different variations of our trained models. Since the transformer predictor and different losses provided minor improvement over the baseline, here we mainly ensemble models with bilstm predictor, pretrained weights, and increase sets of data. 
For ridge regression, we first combine one instance of baseline model and two models trained with de/zh pretrained weights to create 3-dimensional predictions; then adding models with pretrained weights from other 3 languages. After k-fold cross validation, we set our regularization constant as 0.5. 
As shown in Table \ref{tab:results}, the results render slight improvement over the results with the solely baseline model. It is consistent with the improvement of models with pretrained weights while trained individually. 

Next, we apply XGBoost regressor on predictions from more combinations of multiple models including baseline, pretrained weights and various scale of trained data. 
As shown in Table \ref{tab:results}, we obtain overall higher performance for XGBoost ensemble compared to previous weak sub-models. To notice, XGBoost boosts de scores significantly, which intuitively implies the huge prediction disparity between each sub models. Therefore, it is beneficial for the ensemble model to take advantage of better scores disclosed from some of the models based on the sparse weights assigned to them. 
We also observed that XGBoost seems to foster sub-models with pretrained weights over those with more training data. It implies the transfer learning approach generates diverse scores based on individual prior knowledge, which is useful in overall combination. As for sub-models with more data tend to reproduce similar scores with some slightly better improvement, which may not be significantly beneficial for ensembling with similar models. 

Eventually, our best result with all the combination of models both from pretrained weights and more data has Pearson score, 0.298 for de/zh test dataset, which is 2.53 times of original score 0.118 from baseline model. The clear takeway gives that ensembling of all the models can give large gain by exploiting different strength of sub models in some particular nature of data. The results mainly derive from scores with de dataset much more boosted compared to the original baseline model, which is intriguing to further explore the reason of weak performances from sub-models on de dataset.

\begin{table*}
\renewcommand{\arraystretch}{1.3}
\begin{center}
\begin{tabular}{lccccccc}
\hline
\multicolumn{1}{l|}{Model}                           & et     & ne     & ro     & et+ne+ro & de     & zh     & zh+de  \\ \hline \hline
\multicolumn{8}{l}{\textbf{Section \ref{sec:baseline}}: \textit{Baselines}}                                                        \\ \hline
\multicolumn{1}{l|}{RNN predictor + estimator}                        & 0.385  & 0.376  & 0.547  & 0.442    & 0.116 & 0.153 & 0.118 \\ 
\multicolumn{1}{l|}{WMT20 baseline \cite{barrault-etal-2019-findings}}                        & 0.477  & 0.386  & 0.685  & -    & 0.146 & 0.190 & - \\ \hline
\multicolumn{8}{l}{\textbf{Section \ref{sec:transformer}}: \textit{Replace Bilstm by Transformer in predictor}}                                                        \\ \hline
\multicolumn{1}{l|}{Transformer predictor}           & 0.245 & 0.318 & 0.429 & 0.340   & 0.051 & 0.165 & 0.118 \\ \hline
\multicolumn{8}{l}{\textbf{Section \ref{sec:losses}}: \textit{Apply NCE and NEG loss for training Bilstm predictor}}                                                      \\ \hline
\multicolumn{1}{l|}{NCE loss}                        & 0.340   & 0.353  & 0.491  & 0.400      & 0.030   & 0.127  & 0.088  \\
\multicolumn{1}{l|}{NEG loss}                        & 0.353   & 0.367  & 0.498  & \textbf{0.412}      & 0.082   & 0.142  & \textbf{0.108}  \\ \hline
\multicolumn{8}{l}{\textbf{Section \ref{sec:pretraining}}: \textit{Fine-tune pretrained Bilstm predictor with different language pairs}}                               \\ \hline
\multicolumn{1}{l|}{et pretrain}                     & 0.381  & 0.469  & 0.547  & 0.468    & 0.093  & 0.172  & 0.113 \\
\multicolumn{1}{l|}{ne pretrain}                     & 0.388  & 0.388  & 0.554  & \textbf{0.451}    & 0.057  & 0.151  & 0.087  \\
\multicolumn{1}{l|}{ro pretrain}                     & 0.375  & 0.43   & 0.538  & \textbf{0.451}    & 0.105  & 0.212  & \textbf{0.151}  \\
\multicolumn{1}{l|}{de pretrain}                     & 0.379  & 0.407  & 0.581  & 0.469    & 0.104 & 0.204  & 0.131 \\
\multicolumn{1}{l|}{zh pretrain}                     & 0.378  & 0.377  & 0.547  & 0.444    & 0.095  & 0.149  & 0.113 \\ \hline
\multicolumn{8}{l}{\textbf{Section \ref{sec:parallel}}: \textit{Add X de and zh parallel data for training Bilstm predictor}}                                      \\ \hline
\multicolumn{1}{l|}{X=10000 (D1)}                    & 0.407 & 0.422 & 0.529 & 0.455   & 0.068  & 0.195  & \textbf{0.146}  \\
\multicolumn{1}{l|}{X=20000 (D2)}                    & 0.395 & 0.439 & 0.564 & \underline{\textbf{0.473}}   & 0.080   & 0.219  & \textbf{0.146}  \\
\multicolumn{1}{l|}{X=30000 (D3)}                    & 0.357 & 0.425 & 0.579 & 0.467    & 0.111  & 0.185  & 0.142  \\
\multicolumn{1}{l|}{X=50000 (D5)}                    & 0.377 & 0.425  & 0.577 & 0.468   & 0.084  & 0.180   & 0.125  \\ \hline
\multicolumn{8}{l}{\textbf{Section \ref{sec:ensemble_results}}: \textit{Ensemble with ridge regression}}                                                                  \\ \hline
\multicolumn{1}{l|}{base+de+zh}                      & -      & -      & -      & -        & 0.135  & 0.207   & 0.150   \\
\multicolumn{1}{l|}{base+de+zh+ro+et+ne}             & -      & -      & -      & -        & 0.139  & 0.220   & \textbf{0.158}  \\ \hline
\multicolumn{8}{l}{\textbf{Section \ref{sec:ensemble_results}}: \textit{Ensemble with XGBoost regression}}                                                                 \\ \hline
\multicolumn{1}{l|}{base+de+zh+ro}                   & -      & -      & -      & -        & 0.317  & 0.202  & 0.257  \\
\multicolumn{1}{l|}{base+de+zh+ro+et}                & -      & -      & -      & -        & 0.240   & 0.186  & 0.225  \\
\multicolumn{1}{l|}{base+de+zh+ro+et+ne}             & -      & -      & -      & -        & 0.300    & 0.190   & 0.219  \\
\multicolumn{1}{l|}{base+D1}                         & -      & -      & -      & -        & 0.210   & 0.191  & 0.195  \\
\multicolumn{1}{l|}{base+D1+D2+D3}                   & -      & -      & -      & -        & 0.262  & 0.177  & 0.218  \\
\multicolumn{1}{l|}{base+D1+D2+D3+D5}                & -      & -      & -      & -        & 0.270   & 0.194  & 0.240   \\
\multicolumn{1}{l|}{base+D1+D2+D3+de+zh+ro}          & -      & -      & -      & -        & 0.330   & 0.201 & 0.270   \\
\multicolumn{1}{l|}{base+D1+D2+D3+D5+de+zh+ro+et+ne} & -      & -      & -      & -        & 0.386  & 0.223 & \underline{\textbf{0.298}} \\
\hline
\end{tabular}
\end{center}
\caption{QE evaluation results using the Pearson's correlation metric.}
\label{tab:results}
\end{table*}

\section{Conclusions}
In this work, we propose an ensemble-based QE model to enhance QE prediction quality of low-resource translation outputs like English-German and English-Chinese language pairs. We explore several possible settings to incorporate multiple language resources and achieve better performance. Additional parallel language pairs and different losses show the capability to improve the task performance. Our implementations of fine-tuning the pretrained QE models on different languages demonstrate the potential of cross-lingual transfer learning with fewer data. Experimental results also show that our ensemble model outperform the baseline on en-de, en-zh and en-de + en-zh development data.
In future work, we would like to explore on processing of different language pairs and introduce more powerful transfer learning  technique like XLM-R \cite{conneau-etal-2020-unsupervised} on our approaches to gain cross-lingual understanding.




\bibliographystyle{acl_natbib}


\end{document}